\documentclass{article}
\usepackage{ijcai20-multiauthor}

\usepackage{cite}
\usepackage{natbib}

\usepackage{times}

\usepackage{soul}
\usepackage{url}
\usepackage[hidelinks]{hyperref}
\usepackage[utf8]{inputenc}
\usepackage[small]{caption}
\usepackage{graphicx}
\usepackage{amsmath}
\usepackage{booktabs}
\usepackage{comment}
\usepackage{multirow}
\usepackage{balance}
\usepackage{comment}
\usepackage{url}
\usepackage{wrapfig}
\usepackage{changes}
\usepackage{lipsum}
\definechangesauthor[name={Per cusse}, color=orange]{per}
\usepackage{enumitem}
\setlist[itemize]{leftmargin=*,align=left}
\setlist[enumerate]{leftmargin=*,align=left}
\newcommand{\inlinecomment}[1]{}
\usepackage{titling}
\usepackage{cite}

\author{Smit Marvaniya$^1$, Umamaheswari Devi$^1$, Jagabondhu Hazra$^1$, Shashank Mujumdar$^1$, Nitin Gupta$^1$ \\
	$^1$IBM Research - India \\
	{\tt \{smarvani, umamadev, jahazra1, shamujum, ngupta47\}@in.ibm.com}}
\title{\textbf{Small, Sparse, but Substantial: Techniques for Segmenting Small Agricultural Fields Using Sparse Ground Data}}
\inlinecomment{Any Field, Any Place. Any Size, Any Hue: Towards Universal Delineation of Small Agricultural fields with Sparse Ground Truth}
\date{}
\begin{document}
	%
	\maketitle
	\begin{abstract}

		The recent thrust on digital agriculture (DA) has renewed  significant research interest in the automated delineation of agricultural fields. Most prior work addressing this problem have focused on detecting medium to large fields, while there is strong evidence that around 40\% of the fields world-wide and 70\% of the fields in Asia and Africa are small. 
		The lack of adequate labeled images for small fields, huge variations in their color, texture, and shape, and faint boundary lines separating them  make it difficult to develop an end-to-end learning model for detecting such fields. Hence, in this paper, we present a multi-stage approach that uses a combination of machine learning and image processing techniques. In the first stage, we leverage state-of-the-art edge detection algorithms such as holistically-nested edge detection (HED) to extract first-level contours and polygons. In the second stage, we propose image-processing techniques to identify polygons that are non-fields, over-segmentations, or noise and eliminate them. The next stage tackles under-segmentations using a combination of a novel ``cut-point'' based technique and localized second-level edge detection to obtain individual parcels.
		Since a few small, non-cropped but vegetated or constructed pockets  can be interspersed in areas that are predominantly croplands, in the final stage, we train a classifier for identifying each parcel from the previous stage as an agricultural field or not. In an evaluation using high-resolution imagery, we show that our approach has a high F-Score of 0.84 in areas with large fields and reasonable accuracy with an F-Score of 0.73 in areas with small fields, which is encouraging.
		
	\end{abstract}
	
	\section{Introduction}
	\label{sec:intro}
	Digital agriculture (DA) encompasses technologies to enhance the productivity and efficiency of agriculture, thereby bringing critical advantages for farmers and wider social benefits around the world \cite{DA-UN}. A first step in realizing the promise to the farmers is to digitize and create electronic records of their fields. An {\it electronic field record\/} is the ``digital twin'' of a physical field, which can be associated with detailed field information such as boundary, soil type and moisture, pest attacks, crops grown, and yields realized in addition to activities such as pesticide and fertilizer application. By frequently collecting field data and analyzing it, a farmer can be enabled to make more informed decisions. Further, the availability of detailed field activities over the life-cycle of a crop can help  better predict yield quantity and quality, which can  better inform stakeholders downstream, such as commodity traders and financial organizations, to optimize their operations.
	
	Given the urgency of the digitization of field records, several research efforts are underway to detect the boundaries of fields in an automated manner. A comprehensive overview of these efforts is provided in Sec.~\ref{sec:rel-work}. As discussed therein, most of the proposed techniques have been developed and evaluated in the context of large farms, found in countries in the Americas, Europe, and Australia. However, according to ~\cite{lesiv2019estimating}, about 40\% of the fields worldwide are small ranging ~2 hectares in area while more than 70\% (resp. 50\%) of the fields in Asia and Africa are small (resp., very small, ranging less than 0.64 hectares in area). 
	It is in the developing countries in Asia and Africa, such as India and Kenya, that 
	digitization of field records is most lacking and currently involves significant manual effort and hence prone to human and GPS errors, among others. Thus, in these countries, comprehensive and accurate information on fields under cultivation and their status, including field-level agricultural practices, is unavailable.
	
	In this paper, we specifically tackle the problem of delineating the boundaries of small fields, including some very small ones, at \~0.4 ha. Due to their small size and indistinct boundaries delineating them, we use aerial RGB images captured at a high-resolution of 1.19 m. The images we use are DigitalGlobe-based and sourced using Google's Static Maps API.\footnote{\tiny The imagery is accessed from https://maps.googleapis.com/maps/api/staticmap.}
	For our purpose, a field is a single farm-holding with adjacent fields  separated by structures such as a road, a fence, or a very thin strip of uncultivated land. Our objective is to delineate the boundaries and identify each individual field.
	
	Identifying fields from images is a special instance of the image segmentation problem, for which several traditional non-learning based algorithms \cite{Segm:Survey} as well as deep-learning models \cite{DLSegm:Survey} have been developed. In comparison to images of 3-d objects that these methods seek to detect, agricultural fields in aerial images are much simpler. Nevertheless, detecting them, especially small fields, poses  challenges and the existing approaches are not directly applicable. Further, because the fields are small, annotating them to generate sufficient ground truth to train an end-to-end learning model is tedious and time-consuming. Hence, we use a combination of image-processing and machine learning techniques to solve the problem.
	
	\textbf{\hspace{-0.4cm}Contributions.} We present a multi-stage approach for identifying small fields in aerial imagery using limited ground data. We use the state-of-the-art  {\em holistically nested edge detection} (HED) \cite{xie2015holistically} algorithm to initially segment an image into first-level polygons. Since HED is a generic edge-detection network and the open pre-trained model we use has not been trained specifically using features related to crop fields, the output of HED, when applied to field imagery, is a set of  weakly-connected contours containing both false and missing edges. Further, the detected contours are significantly thick, diminishing the size of  detected fields considerably, especially when the fields are small. Hence, we develop and  apply the following novel techniques for culling out final agricultural fields from the output contours of HED.
	\begin{itemize}[leftmargin=*]
		\item  ``Shape-complexity criteria'' for detecting non-fields, over-segmentations, and noise that need to be eliminated.
		\item A method for detecting  {\em cut-points\/}, {\em cuts\/}, and  {\em min-cuts} in non-convex shapes, along which an under-segmented polygon may need to be fragmented.
		\item  Second-level ``localized'' edge detection, tuned to image characteristics local to each first-level  polygon, to identify missing edges in it for further segmentation.
		\item A classifier trained using  shape, color, intensity, and texture-based features of individual parcels to classify the final parcels as agricultural (Ag) or not (non-Ag). 
	\end{itemize}

	\begin{figure*}[t]
		\centering
		\vspace{-10pt}
		\includegraphics[width=\linewidth]{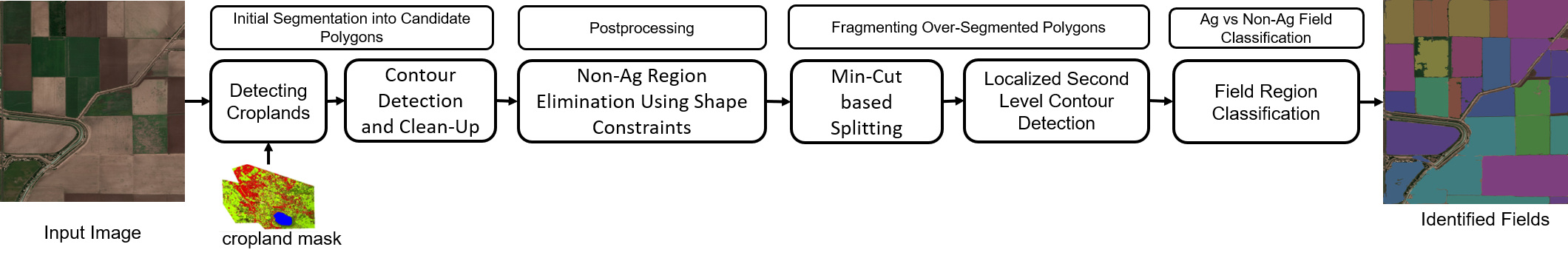}
		\vspace{-20pt}
		\caption{\small Multi-stage framework for field boundary detection. \label{fig:approach}}
		\vspace{-10pt}
	\end{figure*}

	\vspace{-4pt}
	\section{Related Work}
	\label{sec:rel-work}
	Delineating  boundaries and identifying individual fields is a  special instance of the image segmentation problem, with a rich body of literature. The approaches developed therein can broadly be classified into (1) non-learning approaches that use low-level, image-saliency     features~\cite{Segm:Survey} and (2) deep-learning approaches \cite{DLSegm:Survey, badrinarayanan2017segnet, chen2018encoder}. The non-learning approaches can further be classified into edge- or boundary-based, region-based, or hybrid approaches.  Some of these approaches have been re-used and adapted for extracting agricultural fields, as reviewed below.
	
	Work on cropland extraction can  broadly be classified based on the resolution of the images and the spectral information used, the nature of the region it applies to, and the algorithmic techniques used. 
	Early work in this area primarily used remote-sensed satellite imagery with resolution as coarse as 30 m and was hence confined to regions with large fields such as in \cite{EJSB:2002}. In \cite{mueller2004edge}, an approach that combines edge- and region-based methods for identifying fields with boundaries that are typically long with a straight object shape and have high brightness contrast to neighboring regions is presented. The approach presented uses very high resolution (VHR) imagery.
	
	In later work, \cite{yan2014automated} presents another combination of edge- and region-based method, in which, unlike traditional approaches, a time series of Landsat satellite ~\cite{williams2006landsat} images at 30 m resolution is used. The fields considered are large again. 
	\cite{CBGYV:2017} investigates the transferability of {\it gPb\/} contour detection to remotely-sensed
	VHR UAV imagery (of the order of a few cms) and UAV-based cadastral mapping.  \cite{garcia2017machine} solves the problem using ML techniques applied to SLIC  superpixels. Here, the idea is to first over segment the image into superpixels and then train a classifier for each pair of superpixels with a label denoting whether the pair is a part of the same field and can be merged. To tackle sensitivity to intra-plot variability, a subset of authors later explore combining segmentation at different scales using superpixels and multi-temporal images from the same growing season to obtain a single segmentation of the agricultural plots~\cite{GGLR:2018}.  These techniques require high resolution multi-spectral imagery including bands in the non-visible spectrum, which is hard to acquire, and labeled data to train a supervised classifier.
	
	Fueled probably by the growing imperative of DA, there has been a significant surge in the work in this area in the past two years.
	A method for accurately detecting large and mostly rectangular pastures in New Zealand with step edges or linear features over a long distance for boundaries using  time series SPOT satellite imagery is presented in \cite{north2018boundary}. 	 In \cite{XPK:2019}, 
	the authors explore the potential of deep Fully Convolutional Networks (FCNs) for cadastral boundary detection (which is not limited to crop fields) and compare it with traditional segmentation algorithms such as Multi-Resolution Segmentation and gPb in two study sites in Rwanda. To deal with the small field sizes, UAV images with a ground sampling distance of ~2 cm, resulting in VHR imagery, are used. 
	
	Other recent works exploring CNN's include \cite{GLRG:2019, persello2019delineation, XLSY:2018, waldner2019deep}.  In \cite{XLSY:2018}, detecting hard and soft cropland edges from very high resolution (0.8 m) GF-2 using RCF~\cite{RCF:2017} and U-net~\cite{UNet:2015} models is considered.  \cite{GLRG:2019}  uses the U-Net architecture to delineate crop fields and  open data from the Land Parcel Identification System of Spain for training. Images are again of VHR at 25 cm. Results obtained are compared to that of gPb-UCM~\cite{gPbUCM:2011}. \cite{persello2019delineation} uses an FCN-based SegNet architecture to classify boundary pixels and trains it using pan-sharpened multispectral bands of 0.5 m resolution. The fragmented contours are closed to yield closed fields using the OWT-UCM procedure. Finally, \cite{waldner2019deep} uses ResUNet-a, a deep CNN  with a fully connected UNet backbone to compute three values for each pixel: the probability of belonging to a field, the probability of being part of a boundary, and the distance to the closest boundary. The computed values are then used to obtain closed field boundaries.
	
	The recent works described above are quite interesting, but they all require images that have  sub-metre resolution or large-labeled datasets, or have mostly been tested at select regions with large fields, and hence do not suit our purpose.

	\vspace{-2pt}
	\section{Field Boundary Delineation System and Methods}
	\vspace{-5pt}
	\label{sec:fbd-main}
	This section describes the end-to-end stages and steps of our multi-stage pipeline for segmenting field objects. Our input consists of ortho-rectified pan-chromatic aerial or satellite imagery of the larger area of interest.
	The processing steps  have been designed to leverage characteristics unique to field objects to improve the final segmentation. At a high-level, our pipeline consists of four stages as in Fig.~\ref{fig:approach}.
	
	\subsection{Stage 1: Initial Segmentation into Candidate Polygons}
	\label{sec:phase1}
	The first stage consists of (i) a pre-processing step to cull out, at a high-level, parts of an image that correspond to croplands in which to identify fields and (ii) an initial contour detection and segmentation step to identify regions that are candidates for being individual small-holdings, as described below. 
	
	\textbf{\hspace{-0.5cm} Detecting croplands.} In addition to cropland, a larger geography, such as a state or county, typically consists of  non-agricultural land, such as forests, grassland, urban constructed areas, and water bodies. To eliminate processing such regions, the first step of our pipeline masks away non-cropland areas using an open global land cover mask provided by Copernicus~\cite{copernicus-lc-2015} at 100 m resolution.
	
	\textbf{\hspace{-0.4cm}Contour detection.} The next step detects edges within regions identified as croplands. While any edge detection algorithm can be used for this step, in this work, as mentioned in Secs.~\ref{sec:intro} and \ref{sec:rel-work}, we apply the state-of-the-art HED technique as it yielded the best results in our evaluation. HED uses a single stream deep neural network with multiple side outputs that are at different scales.  Though HED can be retrained using aerial field imagery,  due to lack of adequate labeled data, in this work, we use a pre-trained HED network trained on the BSDS500 dataset. 
	As explained in \cite{xie2015holistically},
	\vspace{-0.1cm}	
	\begin{equation*}
	\vspace{-0.1cm}	
	(Y_{\mathrm{fuse}}, Y^{1}_{\mathrm{side}}, ..., Y^{N}_{\mathrm{side}}) = \mathrm{CNN}(I^{\mathrm{input}}, (W, w, h)^{*}),
	\end{equation*}
	where $\mathrm{CNN}(\cdot)$ represents the edge map produced by the HED network, $Y^{i}_{\mathrm{side}}$ denote the various side outputs, and $Y_{\mathrm{fuse}}$, the weighted fusion output of the side outputs. $W$, $w$, and $h$  denote  the standard network layer parameters, weights for side output layers, and fusion weights (for combining the side outputs), resp., which are learned as a part of the pre-trained model.  The final edge map is obtained by further combining the edge maps generated at multiple scales by the side outputs as
	$Y_{\mathrm{hed}} = \mathrm{Average}(Y_{\mathrm{fuse}}, Y^{1}_{\mathrm{side}}, ..., Y^{N}_{\mathrm{side}}).$
	
	We clean-up the edge map generated by HED by applying image operations such as erosion, dilation, and explicit edge-thinning. From the cleaned-up edge map, an initial set of polygons, which are candidates for being fields, are extracted by performing a connected-components extraction step.
	
	The candidate polygons obtained  require further processing due to the following reasons: (i) Some polygons may not correspond to crop fields but could be isolated buildings or built-up areas or non-crop but pockets with natural vegetation that are interspersed within regions that are predominantly croplands, and hence should not be classified as crop  fields (or Ag fields), 
	(ii) the polygons could be under-segmented fields, wherein multiple adjacent fields are agglomerated, or (iii) the polygons could be over-segmented wherein a single field is split into multiple pieces. In what follows, we describe  the techniques we use to address the above issues.
	When the fields are small, under-segmentation is more predominant than over-segmentation, and as such, we focus more on handling the former.
	
	\subsection{Stage 2: Shape-based non-Ag and Spurious Field 
		Elimination}
	\label{sec:scc}
	
	
	We identify non-Ag regions using two different methods in two stages. We now describe the first method, which deals with small spurious polygons and noise. Method~2 is applied in Stage~4 after all significant parcels have been detected. 
	
	The first method consists of a set of heuristics (listed below) that leverage the layout characteristics of crop fields, which are typically convex, and have a minimum size and aspect ratio (ratio or width to length) based on the geography.

\begin{figure}[t]
		\centering
		\includegraphics[width=0.7\linewidth]{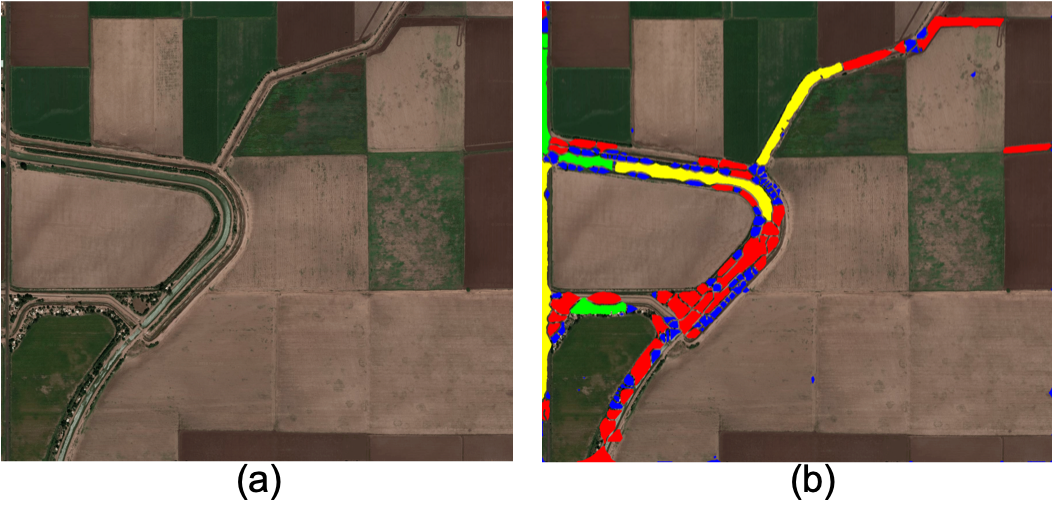}
		\vspace{-10pt}
		\caption{\small Sample image with spurious fields identified using shape  constraints of Sec.~\ref{sec:scc}. (a) Input image. (b) Spurious fields identified via the four conditions of Sec.~\ref{sec:scc} marked in order using blue, green, red, and yellow colors. \label{fig:shape_complexity_constraints}}
		\vspace{-12pt}
	\end{figure}

	\begin{enumerate}
		\item \textbf{Very small, noisy polygons removal}:  Both the perimeter and area of a field are typically reasonably large, with the threshold depending on the geographic location. Hence, the first rule drops all polygons with perimeter and area less than specified thresholds.

		\item \textbf{Convexity threshold}:  The next condition ensures that the contour of the detected polygon does not deviate much from being convex. (Perfect convexity cannot be expected because of some irregularities in the edges.) For this, for polygons with area below a specified threshold, we verify that the ratio between the areas of the convex hull drawn around the polygon and the polygon per se. is not large.\footnote{Polygons that are significantly non-convex but larger in area than the threshold are addressed by the sub-segmentation technique discussed in Sec. \ref{sec:splitting}}
		\item \textbf{Small noisy, elongated polygons removal:} To eliminate small polygons that pass the above two tests, but are still noisy, we require that for each polygon with area below a specified minimum, the ratio between its area and perimeter exceeds a threshold. This condition caters to removing within-field over-segmentations due to small constructions, prominent water channels, etc.

		\item \textbf{Long, thin strips removal:} A long, thin strip with a regular shape, {\it e.g.,\/} a rectangle, can be easily identified by its aspect ratio. On the other hand, identifying strips with irregular shapes without well-defined length or width, as shown in yellow in Fig.~\ref{fig:shape_complexity_constraints}(b), is a bit tricky. Such long thin strips are mostly non-croplands and can correspond to footpaths in fields, roads, long waterways, etc. To detect such strips, the width of a polygon is computed at each of its boundary pixels as the minimum of the horizontal and vertical distances (along the $x$ and $y$ axes, resp.) spanned by the polygon at that pixel. 
		The width of the polygon is then given by the average width at all the boundary pixels.
		The approximate length of the polygon can then be obtained using its average width and  perimeter, from which its aspect ratio can be computed.
		Polygons with aspect ratio below a specified threshold are then discarded.
	\end{enumerate}
	
	Fig.~\ref{fig:shape_complexity_constraints} shows a sample image with spurious fields identified using the above constraints.
	\subsection{Stage 3: Splitting Under-Segmented Polygons}
	\label{sec:splitting}
	In regions with small fields,
	first-level polygons could be grossly under-segmented with several missing edges due to indistinct boundaries, noise, coarse image resolution, etc.. Further, the edge detector we use, HED, is trained on a generic set of images and oblivious to the distinct aspects of fields and their bounding structures. Hence, in this stage, we identify and fragment under-segmented polygons.
	\vspace{-5pt}
	\subsubsection{Splitting via Min-Cuts} The first method splits large polygons via high-curvature analysis of its contour.  Points on the contour are analyzed to identify {\it cut points\/}, from which {\it min-cuts} -- lines along which to split the polygon -- are determined.
	
\begin{figure}[t]
	\includegraphics[width=\linewidth]{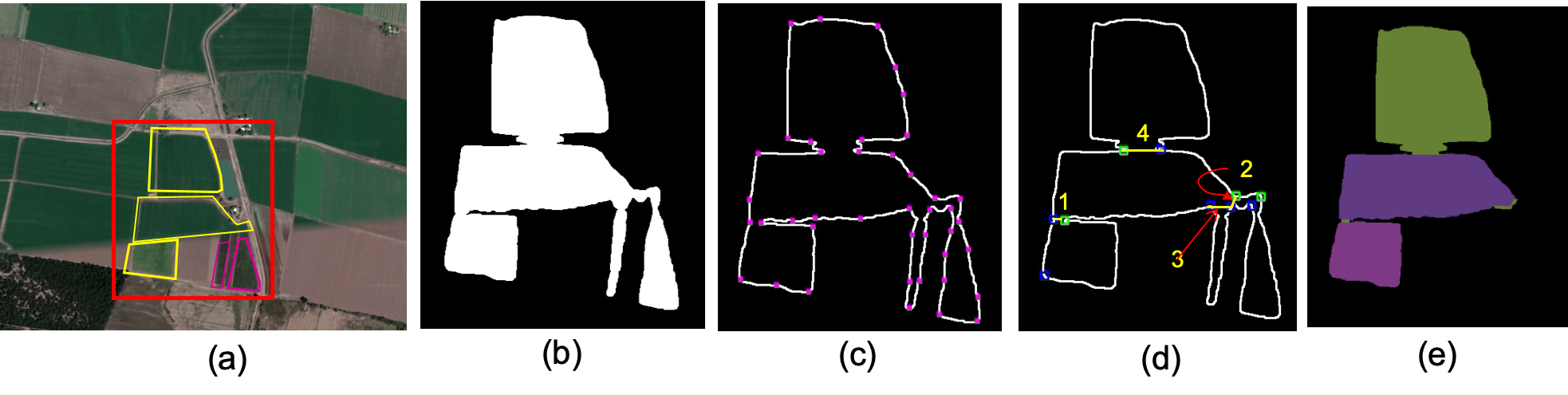}
	\vspace{-20pt}
	\caption{Splitting under-segmented fields using min-cuts. (a) Input Image with Ag fields (yellow) and non-Ag fields (purple) of interest. (b) Under-segmented polygon agglomerating multiple Ag fields and non-Ag polygons. (c) Identified high curvature points. (d) Initial cut points, candidate cuts,  and min-cuts (yellow lines) in the order selected. (e) Splitting at the min-cuts and discarding noisy fragments for the final Ag set. \label{fig:cut_point_steps}}
	\vspace{-12pt}
\end{figure}

	Many a time, faint boundaries separating fields are (i) only partly detected by edge detection algorithms resulting in distinct fields that are held together by short strips  or (ii) largely undetected, leaving the fields mostly connected, leading to  odd, non-convex polygons. (See Fig. ~\ref{fig:cut_point_steps}(b).) The purpose of this step is to identify such instances of missing edges and fragment the polygons along those.
	
	Curvature on the contour at one or both the endpoints of missing edges in cases as described above will, in general, be quite high. So, to detect the missing edges, we first detect high-curvature points on the contour, which we call {\em cut points\/}. Each cut point $c$ is paired with another point on the contour
	to which its euclidean distance is the shortest and the contour distance to which is greater than the euclidean distance to it. Each such pair is termed a {\em cut\/}.  Cuts for which the euclidean distance between its end points is less than a threshold whereas the contour distance between the same points is greater than a second threshold are chosen as the set of candidate cuts. (Denoted by blue-yellow pairs in Fig.~\ref{fig:cut_point_steps}(d).) From this set, the final min-cuts are culled out recursively as follows. The initial min-cut selected is the one with maximum value for $\alpha_{\mathrm{cut}(p_i,p_j)}\cdot \mathcal{C}_s(p_i,p_j)$, where (i) $p_i$ and $p_j$ are the end points of  $\mathrm{cut}(p_i,p_j)$, (ii) $\alpha_{\mathrm{cut}(p_i,p_j)}$ is a binary flag that is set to 1 if the line between $p_i$ and $p_j$ is within the polygon and the contour lengths of the two sub-polygons produced by the cut  exceed a threshold, and (iii) $\mathcal{C}_s$ measures the strength of edge $(p_i,p_j)$ using a combination of edge distance metric $C_{\mathrm{dist}}$ and edge probability metric $\mathcal{C}_{\mathrm{prob}}$ as $\mathcal{C}_{s}(\mathbf{p_i}, \mathbf{p_{j}}) = \beta\cdot\mathcal{C}_{\mathrm{dist}}(\mathbf{p_i}, \mathbf{p_{j}}) + (1-\beta)\cdot\mathcal{C}_{\mathrm{prob}}(\mathbf{p_i}, \mathbf{p_{j}})$, where $0 \leq \beta \leq 1$. $\mathcal{C}_{dist}$ is computed by measuring the distance between the cut segment and the contour map using directional chamfer distance \cite{liu2010fast} whereas $\mathcal{C}_{prob}$ is computed by estimating the normalized edge probability along the cut. 

We recursively apply the above procedure to each of the sub-polygons produced by the initial min-cut to identify additional min-cuts to split the sub-polygons. We terminate when there are no more qualifying cuts. 

Fig.~\ref{fig:cut_point_steps} shows an example  in which multiple Ag and non-Ag fields are agglomerated into a single polygon and illustrates the steps involved in appropriately splitting them using min-cuts. 

\subsubsection{Splitting via Localized Second-Level Contour Detection}

The performance of edge-detection algorithms is impacted by the heterogeneity of color, texture etc. within an image, and if based on a supervised model, on the differences between the images in the training and test sets.  Most algorithms are dependent on a number of configurable parameters for effectiveness, which are typically based on the mean values of the pixels in the training or test images or both. This kind of averaging manifests as under-segmentation, as the edges are sharper only when the intensity gradient is large.
\vspace{-0.2cm}
\begin{wrapfigure}{r}{4.2cm}
	\centering
	\includegraphics[width=1.1\linewidth]{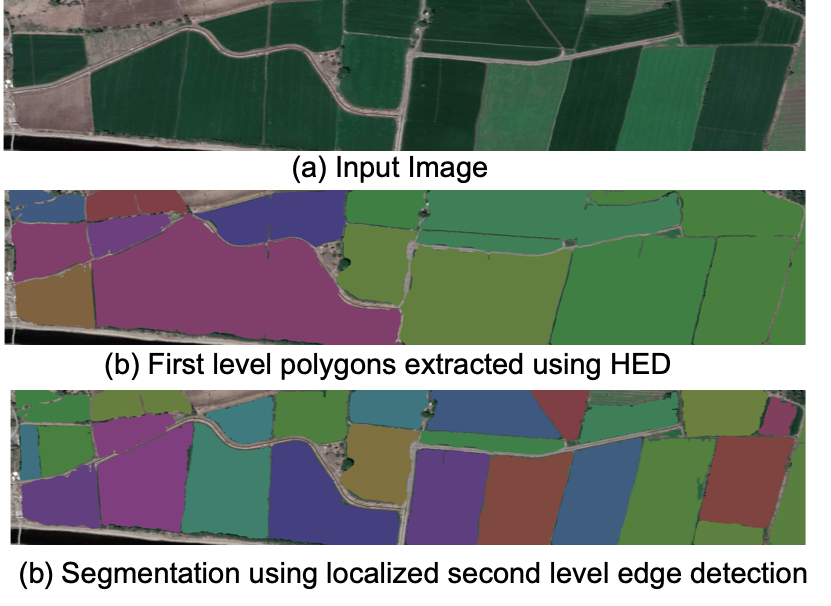}
	\vspace{-18pt}
	\caption{\small Sample localized second-level contour detection. \label{fig:second_level_contour_detection}}
	\vspace{-12pt}
\end{wrapfigure}

We deal with heterogeneous landscapes by performing a second-level ``localized'' contour detection separately on each of the polygons detected at the end of the previous step. The intuition is that configuring the edge detection algorithms using local means (of the individual polygons) should help in better detecting less distinct edges present within homogeneous regions. 
This indeed turns out to be the case. Fig.~ \ref{fig:second_level_contour_detection} shows an example of localized second-level edge detection using canny edge detector~\cite{cannyedge}, wherein hysteresis thresholds for the algorithm are based on the mean of pixels in the polygon under consideration, as opposed to the entire image.

\subsection{Stage 4: Ag vs. Non-Ag Classification}
\label{sec:ag_nonag}
As discussed in Sec.~\ref{sec:intro}, land cover classification maps are of coarse resolution of 100 m or worse, and hence cannot be used to detect small pockets of constructions or natural vegetation interspersed in predominantly agricultural regions. The parcels identified in the previous stage can hence be either crop fields or others. In this final stage we identify parcels that are crop-based. 

The spectral and shape features of agricultural (Ag) fields are considerably different from the others (non-Ag). E.g., Ag fields have a reasonably definite shape, which is mostly convex with only a few vertices, and distinct shade and texture. Thus, we build a training set consisting of Ag and non-Ag parcel images for each of which we  extract its (i) shape features such as  perimeter and area, convex hull perimeter and area, (ii) color distribution in the form of color histogram computed using \cite{datta2006studying}, and (iii) texture histogram, capturing the distribution of {\it local binary patterns\/} \cite{lbp} of pixels in the parcels (computed using a radius of 3). 

The above features are then used to train a random forest model for classifying a detected parcel as Ag or non-Ag. Each resultant parcel at the end of the previous stage is classified using this model, yielding the final set of Ag fields.

\vspace{-0.2cm}
\section{Empirical Evaluation}
\label{sec:experiments}
\vspace{-0.2cm}
This section evaluates our approach and  presents its results.
\vspace{-0.2cm}
\subsection{Datasets and Experimental Setup}
\vspace{-0.1cm}
We evaluate our approach on two different datasets given below covering multiple geographical locations.  (Though our objective is to effectively delineate small farms, we evaluate it on regions with larger farms, too, to show that our approach is generic and  extensible with minimal tuning.)
\begin{enumerate}
	\item Large Field Region (USA): Contains 10 images from different regions of the United States. The average area captured by each image is around 5 sq. km. and the average area of each field is around 11 hectares. The images together span a total area of ~50 sq. km. 
	\item Small Field Region (India and Kenya): Contains 15 images from different regions of developing countries such as India and Kenya. In this case, the images are smaller by a factor of 5 at 1 sq. km. while the fields are smaller by a factor of about 10 at 1 hectare. About 50\% of the fields are very small ranging less than 0.5 hectares in area. The images span around 15 sq. km.
\end{enumerate}

We used the online image annotation tool \emph{ImgLab}\footnote{https://github.com/davisking/dlib/tree/master/tools/imglab} developed by MIT to manually mark boundaries of fields in each of the images in the dataset to generate the ground truth. 


\begin{table}
	\tiny
	\centering
	\begin{tabular}{|l|c|c|c||c|c|c|}
		\hline
		\multicolumn{1}{|c|}{\multirow{2}{*}{\textbf{Stage}}} & \multicolumn{3}{c||}{\textbf{Large Field Region (USA)}}                                                                               & \multicolumn{3}{c|}{\textbf{Small Field Region (India)}}                                                                            \\ \cline{2-7} 
		\multicolumn{1}{|c|}{}                       & \multicolumn{1}{l|}{\textbf{Precision}} & \multicolumn{1}{l|}{\textbf{Recall}} & \multicolumn{1}{l||}{\textbf{F1}} & \multicolumn{1}{l|}{\textbf{Precision}} & \multicolumn{1}{l|}{\textbf{Recall}} & \multicolumn{1}{l|}{\textbf{F1}} \\ \hline \hline
		\textbf{PP}                                & 0.6552                              & \textbf{0.9536}                           & 0.7675                       & 0.5615                              & \textbf{0.8751}                           & 0.6787                       \\ \hline
		\textbf{PP+MC}                           & 0.6836                              & 0.9521                           & 0.7812                       & 0.5979                              & 0.8716                           & 0.7047                       \\ \hline
		\textbf{PP+LCD}                          &  0.8721                                   &     0.8394                            &   0.8463                           & 0.6060                               & 0.8679                           & 0.7095                       \\ \hline
		\textbf{PP+MC+LCD}                     & \textbf{0.8710}                               & 0.8409                           & \textbf{0.8467}              & \textbf{0.6858}                              & 0.7912                           & \textbf{0.7290}               \\ \hline
	\end{tabular}
	\caption{\small Ablation study of  the multi-stage approach on large  and small fields. PP: non-Ag fields and noise removal using post-processing step, MC: Min-cuts, LCD: Localized level-2 contour detection. \label{ablation_study}}
	\vspace{-15pt}
\end{table}
\subsection{Results}
\label{sec:results}

\begin{figure*}[h]
	\includegraphics[width=0.95\linewidth]{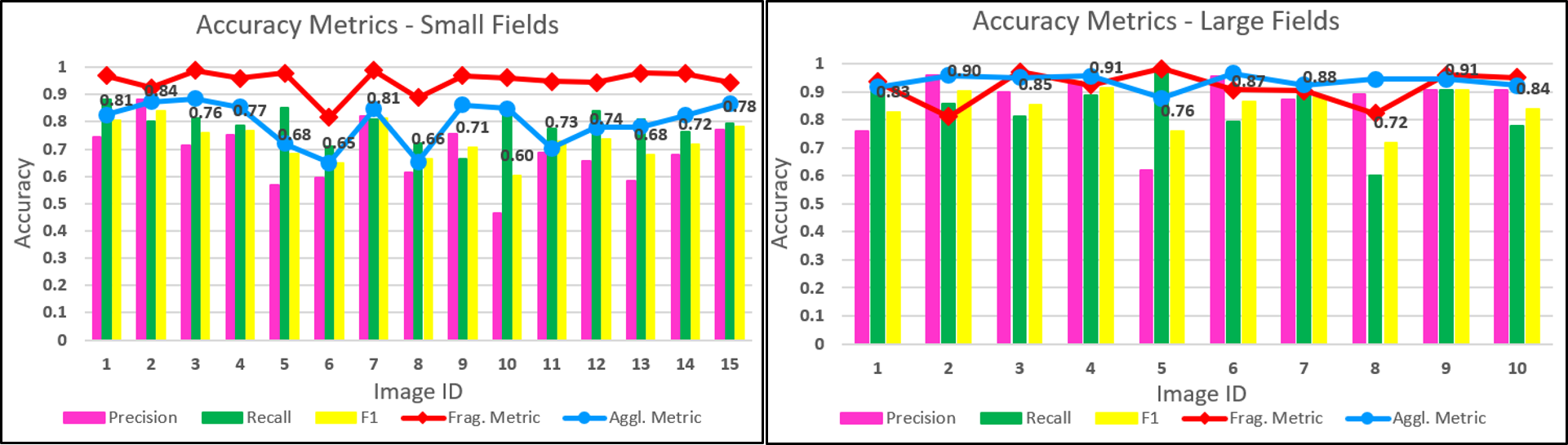}
	\caption{\small Accuracy metrics for small (India) and large (USA) fields. The numbers on the chart denote F1 scores. \label{fig:metrics-main}}
	\vspace{-10pt}
\end{figure*}

\begin{figure*}
	\centering
	\includegraphics[width=\linewidth]{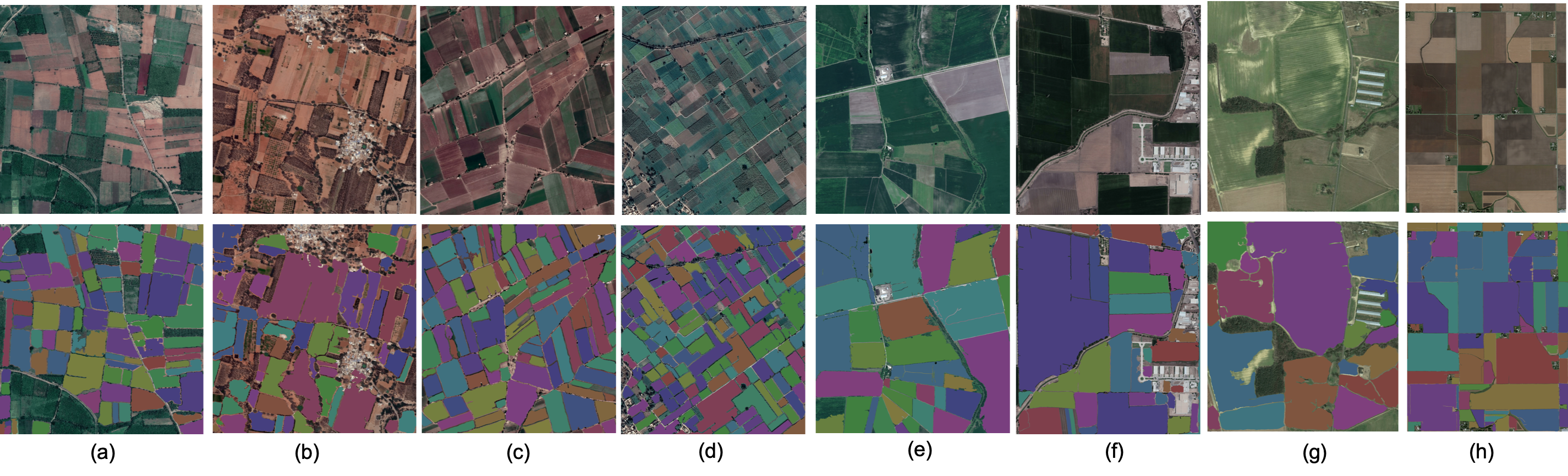}
	\vspace{-22pt}
	\caption{\small Sample visual results for field boundary delineation using the multi-stage approach for {\bf(a)--(d)} small fields and {\bf(e)--(h)} large fields. Top line has vanila images while the bottom line is with overlaid fields. \label{fig:visual_results}}
	\vspace{-10pt}
\end{figure*}

\subsubsection{Overall boundary detection} To compute the accuracy metrics, we generate a mapping from the set of fields in the ground truth to the fields in the detected output. Our mapping effectively partitions the union of the two sets into three subsets: the first subset identifies instances of over-segmentation and hence each ground truth field in this set is mapped to more than one detected field, the second does the inverse of identifying instances of under-segmentation and hence each detected field in this set is mapped to multiple ground truth fields, while the third subset contains fields that are rightly segmented and hence has a one-to-one mapping from the ground truth subset to the detected subset. It is possible for a ground-truth or detected field to be part of multiple under- or over- or right segmentation instances or a combination. In such cases, a field is assigned to the instance in which there is  maximal overlap between the ground truth and detected fields.

With the fields partitioned as above, accuracy numbers are generated for each mapping instance. Pixels that belong to both the ground truth and detected fields are labeled true positive pixels and those that belong only to a ground truth (resp., detected) fields as false negative (resp., false positive) pixels, using  which precision, recall, and F-score (micro F1) are generated for each mapping instance. The ratio of the number of ground truth (resp., detected) fields mapped to the detected (resp., ground truth) fields in each instance provides the agglomeration (resp., fragmentation) metric. If $k$ is the number of sub fields in an agglomeration or fragmentation, then the corresponding metric is given by $\frac{1}{1+\log(k)}$.
The numbers are averaged over all the mapping instances weighted by the total number of ground truth pixels in each instance to arrive at per-image accuracy metrics.  

Image-level metrics are shown in Fig.~\ref{fig:metrics-main}. Overall, we detect more than 98\% of the large fields and 95\% of the small fields. 
The average F1 score for the detected large fields is around 84.5\% while it is ~73\% for small fields (see Table~\ref{ablation_study}). 
Fragmentations are minimal for both large and small fields, with 0.92 and 0.94, resp., as averages. On the other hand, agglomerations are higher for small fields with 0.79 as the average metric. This is due to the faint boundaries between small fields and is as expected.
Given the ambiguous nature of the small-field images, the results are quite promising. Qualitative visual results are shown in Figs. \ref{fig:visual_results}(a)--(d) for small fields and Figs.~\ref{fig:visual_results} (e)--(h) for large fields.

\begin{table}
	\small
	\centering
	\resizebox{0.9\linewidth}{!}{
		\begin{tabular}{|c|c||c|c|c|}
			\hline
			\multicolumn{2}{|c|}{} &\multicolumn{2}{|c|}{Predicted}\\\cline{3-4}
			\multicolumn{2}{|c|}{} & \multicolumn{1}{|c|}{Ag Fields} & \multicolumn{1}{|c|}{Non Ag Fields}  \\\hline\hline
			\multirow{2}{*}{Ground Truth}& Ag Fields & 179    & 3\\\cline{2-4}
			& Non Ag Fields & 5    & 25\\\hline
			\multicolumn{4}{|c|}{Macro-F1: 92\%, Accuracy: 96.22\%}\\\hline
		\end{tabular}
	}
	\vspace{-8pt}
	\caption{\small Confusion matrix for Ag vs non-Ag field classification.}
	\vspace{-8pt}
	\label{tab:fbi_ag_vs_non_ag}
\end{table}

\vspace{-0.2cm}
\subsubsection{Ag vs. Non-Ag Classification}
\label{sec:NONAG-res}
We also assessed the accuracy of the random forest classifier that we use in the  final stage for culling out agricultural parcels. For this, we build a dataset consisting of around 1025 fields captured in a small set of 25 images (which are orthogonal to the set used above). The images are run through our pipeline to delineate the parcels and also extract the needed features (see Sec.~\ref{sec:ag_nonag}).  We manually label each identified parcel as Ag or non-Ag and perform 5-fold cross-validation using the extracted features. Accuracy numbers are reported in Table~\ref{tab:fbi_ag_vs_non_ag}.

\vspace{-0.1cm}
\subsection{Ablation Study}
We also performed an ablation study to assess the relative importance and contribution of some of the important steps of the multi-stage approach. The steps we evaluate are post-processing in Stage~2 to remove small non-Ag fields and noise (PP), splitting under-segmented fields using min-cuts (MC) and localized level-2 contour detection (LCD). Accordingly, we run the pipeline with PP, PP+MC, PP+LCD, with the final Ag vs. non-Ag (NONAG) enabled in each case. NONAG is evaluated separately (in Sec.~\ref{sec:NONAG-res} above). Numerical results are provided in Table~\ref{ablation_study}. Our key observations are as follows:
\begin{itemize}
	\item PP is able to identify most of the noisy fields and has very high recall of 0.95 and 0.88 for large and small fields, respectively.
	\item Adding min-cuts based splitting of under-segmented fields (MC) improves the precision and F1 for both the datasets without adversely impacting recall.
	\item Performing localized level-2 contour detection (LCD) to handle under-segmentation improves both precision and F1 significantly for both large and small fields.
	\item Precision and F1 improve substantially by 8 and 5 points for large and small fields, resp., when MC and LCD are combined, in comparison to terminating with PP in Stage~2.
\end{itemize}

\vspace{-0.2cm}
\section{Discussion \& Conclusion}
\label{sec:conclusion}
\vspace{-0.2cm}
In this paper, we addressed the problem of accurately segmenting small  fields with indistinct boundaries, which dominate the agricultural landscape in developing countries and constitute a substantial chunk globally. Unlike most other objects, field objects lack a well-defined shape and can assume varying shades of color and texture. Further, there is also a paucity of labeled ground truth that can adequately represent small fields world-wide. Due to these reasons, building an end-to-end learning model for segmenting small fields is not effective, and we instead developed a multi-stage framework, in which we leverage generic image-segmentation techniques and devise novel domain-specific techniques. Our approach only required a small set of 25 minimally annotated images for training in the final stage as opposed to thousands that would be required for an end-to-end learning-based approach.

In an empirical evaluation over fields spread across widely differing geographies, the accuracy of our approach is promising.  There is also scope to enhance the approach using, for instance, multi-temporal data, which we plan to consider in future work. 
Further, the work in this paper can be used to generate much-needed ground truth for seeding end-to-end data-driven learning and bootstrap the creation of a classifier based on it.
\bibliographystyle{ijcai20-multiauthor}
\bibliography{ijcai20-multiauthor}
\end{document}